\documentclass[letterpaper, 10 pt, conference]{ieeeconf}  
\IEEEoverridecommandlockouts 

\usepackage{amssymb}
\usepackage{tensor}
\usepackage{cancel}
\usepackage{algorithm}
\usepackage{mathtools}

\usepackage{tabularx}
\usepackage[table]{xcolor}
\usepackage{booktabs} 
\usepackage{makecell}
\usepackage{multirow}

\usepackage{float}
\usepackage{graphicx}
\usepackage{subfig}

\usepackage{xparse} 
\usepackage{xcolor}
\usepackage{framed} 
\usepackage{hyperref}
\usepackage[acronym]{glossaries}
\usepackage{pifont} 
\usepackage{lipsum} 

\newcommand{\figref}[1]{Fig.~\ref{#1}}
\newcommand{\secref}[1]{Sec.~\ref{#1}}

\NewDocumentCommand{\tf}{O{} O{} m o O{}}  
  {
    \tensor*[^{#1}_{#2}]{\mathbf{#3}}{_{#4}^{#5}}
  }

\providecommand{\R}{\mathbb{R}} 

\begin{document}

\title{\LARGE \bf Raising Body Ownership in End-to-End Visuomotor Policy Learning via Robot-Centric Pooling}
\author{Zheyu Zhuang$^{1}$ Ville Kyrki$^{2}$ Danica Kragic$^{1}$
\thanks{$^{1}$The authors are with the Robotics, Perception and Learning Lab, EECS, at KTH Royal Institute of Technology, Stockholm, Sweden
     {\tt\small zheyuzh, dani@kth.se}}%
\thanks{$^{2}$The authors are with the Intelligent Robotics Group, Department of Electrical Engineering and Automation (EEA), Aalto University, Espoo, Finland.
     {\tt\small ville.kyrki@aalto.fi}}%
}

\maketitle

\begin{abstract}
    We present Robot-centric Pooling (RcP), a novel pooling method designed to enhance end-to-end visuomotor policies by enabling differentiation between the robots and similar entities or their surroundings.
Given an image-proprioception pair, RcP guides the aggregation of image features by highlighting image regions correlating with the robot's proprioceptive states, thereby extracting robot-centric image representations for policy learning.
Leveraging contrastive learning techniques, RcP integrates seamlessly with existing visuomotor policy learning frameworks and is trained jointly with the policy using the same dataset, requiring no extra data collection involving self-distractors. 
We evaluate the proposed method with reaching tasks
in both simulated and real-world settings.
The results demonstrate that RcP significantly enhances the policies' robustness against various unseen distractors, including self-distractors, positioned at different locations.
Additionally, the inherent robot-centric characteristic of RcP enables the learnt policy to be far more resilient to aggressive pixel shifts compared to the baselines.

\end{abstract}

\hypersetup{
    colorlinks=true,
    linkcolor=blue,       
    citecolor=blue,       
    filecolor=blue,       
    urlcolor=blue         
}





\section{Introduction}

Body ownership enables us to differentiate our own body from objects in our surroundings (\textit{self-recognition}) and from other individuals (\textit{self-other distinction})~\cite{kilteni_2015_MyFakeBody}.
These aspects are also crucial for robots, especially in shared or multi-robot settings, where a robot’s actions should remain unaffected by environmental objects and other robots. Acknowledging that body ownership in humans involves complex multisensory integration and cognitive processes~\cite{synofzik2013experience}, and drawing on prior work~\cite{lanillos2020robot_self_other, yang2020morphology, toshimitsu2022_dije_self_recognition}, we describe body ownership for robots at the visuomotor level, focusing on simulating features of self-recognition and self-other distinction.

\begin{figure}
\centering
    \subfloat[Testing Sample]{\includegraphics[width=0.15\textwidth]{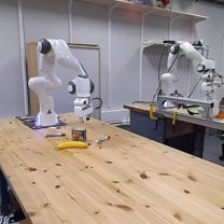}\label{subfig:train_im}}%
    \hfill
    \subfloat[IPA Scores $\vert\, \boldsymbol{p^+}$]{\includegraphics[width=0.145\textwidth]{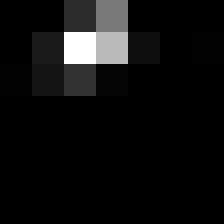}}%
    \hfill
    \subfloat[Saliency $\vert\, \boldsymbol{p^+}$]
    {\includegraphics[width=0.145\textwidth]{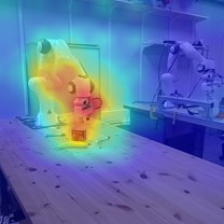}}%
    \\
    \subfloat[IPA Scores $\vert\, \boldsymbol{p^-}$]{\includegraphics[width=0.145\textwidth]{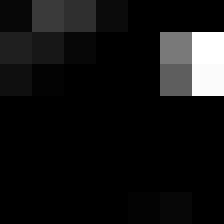}}%
    \hfill
    \subfloat[Saliency $\vert\, \boldsymbol{p^-}$]
    {\includegraphics[width=0.145\textwidth]{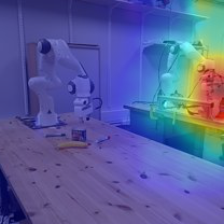}}%
    \hfill
    \subfloat[Saliency $\vert$ \textbf{Baseline}]{\includegraphics[width=0.145\textwidth]{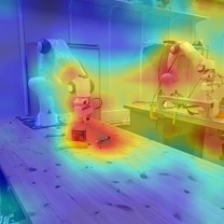}}%
    \caption{\textbf{Body ownership via Robot-centric Pooling.} RcP enables a conventional policy regression baseline to foster self-recognition and the ability to distinguish self from others. \textbf{(a):} A testing sample including a self-distractor (right). \textbf{(b):} Image-proprioception alignment scores for self-state, $\boldsymbol{p^+}$. \textbf{(c:)} Image saliency map~\cite{srinivas2019fullgrad} with $\boldsymbol{p^+}$ based on regressed policy (warmer colours indicate higher relevance). \textbf{(d):} IPA scores for the distractor's state $\boldsymbol{p^-}$. \textbf{(e):} Saliency map with $\boldsymbol{p^-}$. \textbf{(f):} Saliency map from the Spatial-Softmax~\cite{finn2016_spatial_encoder_vs} baseline. }
    \label{fig:fullgrad_1}
    \vspace{-2mm}
\end{figure}

In this work, we introduce Robot-centric Pooling (RcP) to address severely limited self-recognition and self-other distinction capability in conventional end-to-end visuomotor policy learning.
This novel pooling method explicitly integrates both visual information and the robot's proprioceptive state, setting it apart from traditional pooling methods that rely solely on image data.
RcP computes alignment scores between image regions and the robot's proprioceptive states to derive image representations that reflect a robot-centric perspective (Fig.\ref{fig:fullgrad_1}).
Notably, RcP is fully self-supervised and task-agnostic, allowing it to integrate seamlessly with existing standard CNN-based visuomotor policy learning frameworks.
It can be jointly trained with the policy using the same training data, requiring no additional data collection.

\begin{figure*}
    \centering
    \subfloat[System Overview]{
    \includegraphics[height=0.21\textwidth]{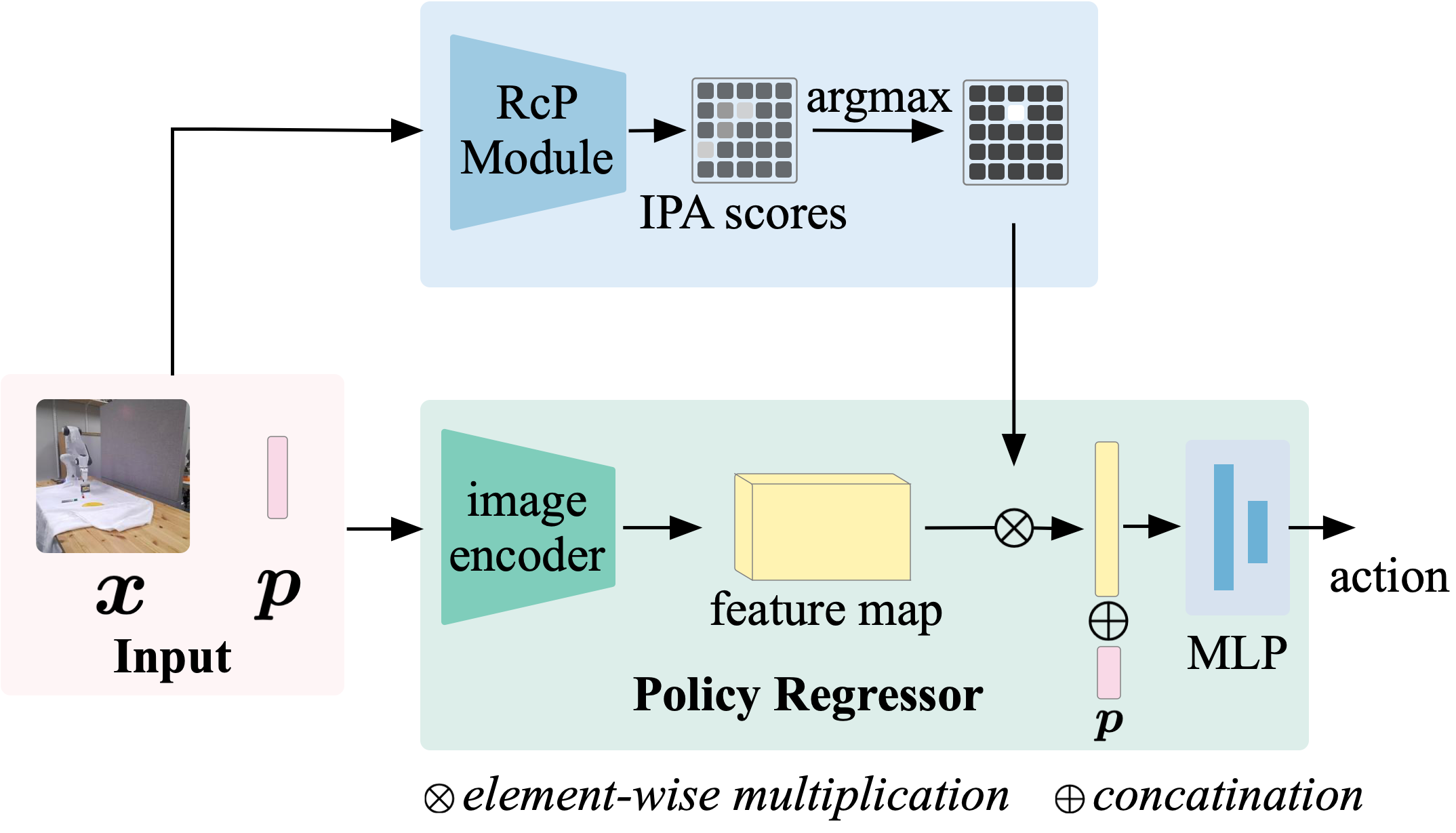}\label{subfig:system_overview}
    }
    \hspace{1cm}
    \subfloat[Robot-centric Pooling Module]{\includegraphics[height=0.21\textwidth]{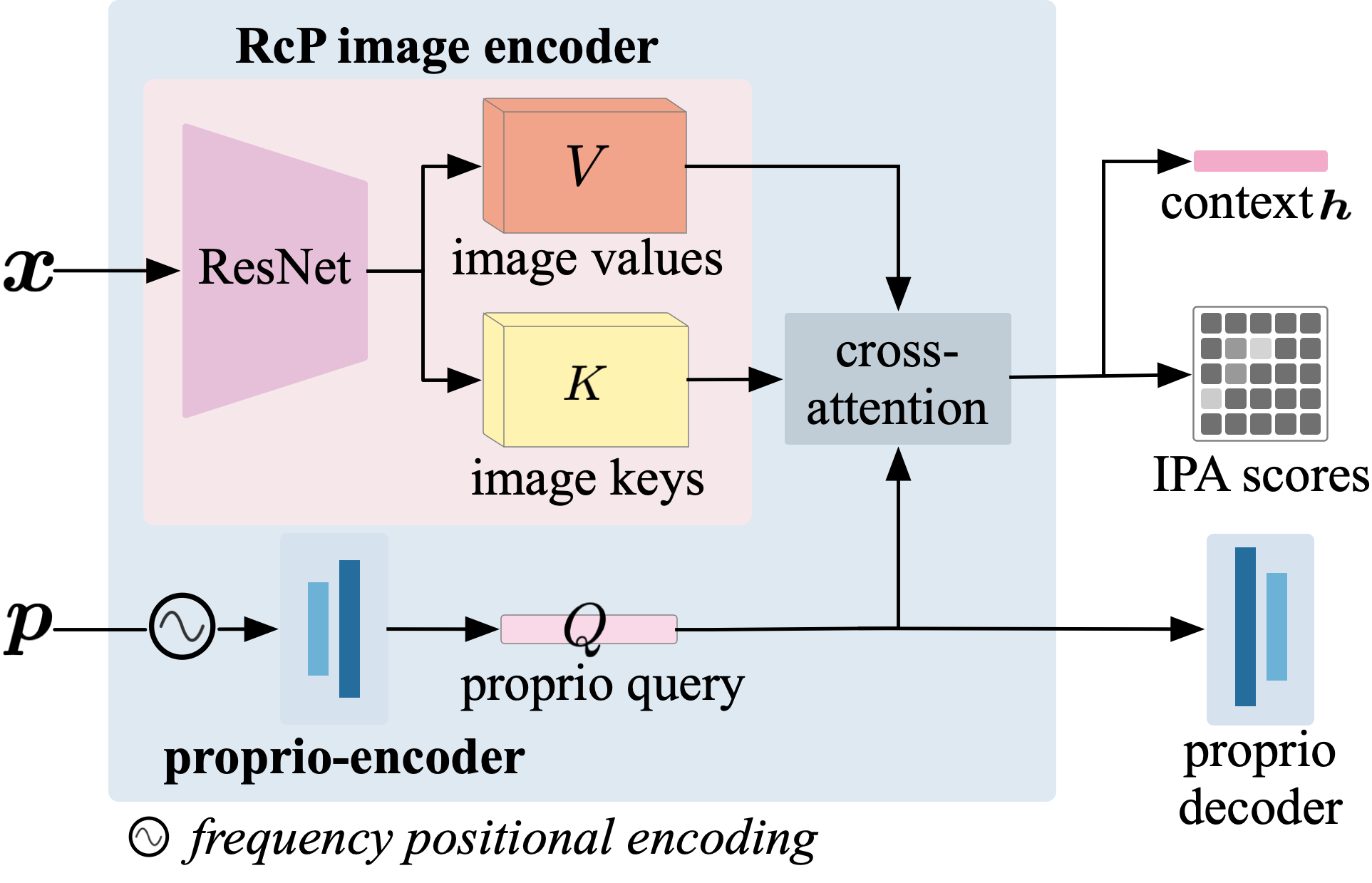} \label{subfig:RcP}}
    \caption{\textbf{System Overview and the Robot-centric Pooling Module.}
    \textbf{(a):} Robot-centric Pooling extracts the most relevant feature corresponding to the identified self for the regression task.
    \textbf{(b):} RcP computes Image-Proprioception Alignment (IPA) scores from an image-proprioception pair $(\boldsymbol{x}, \boldsymbol{p})$ and aggregates image values accordingly to create a context vector for contrastive learning and image representation in the regression pipeline.
    }
    \vspace{-2mm}
\end{figure*}

We evaluate RcP with reaching tasks,
in both simulated and real-world settings. 
Our experimental results show that:
\begin{itemize}
\item Conventional end-to-end learning baselines exhibit considerable sensitivity to environmental distractions and self-distractors, revealing a fundamental deficiency in the development of body ownership.
\item Robot-centric Pooling (RcP) demonstrates significant enhancement against distractions, showing only a slight decrease in success rates (from 96\% to 92\%) amidst a self-distractor in real-world experiments, as opposed to baseline models which plummet to below 15\%.
\item Benefiting from the robot-centric nature, RcP significantly enhances policy robustness against aggressive image shifts compared to baseline methods.
\end{itemize}
To the best of our knowledge, this is the first demonstration of both the self-other distinction and the self-recognition capabilities of body ownership in the context of end-to-end visuomotor policy learning.

\section{Related Work}
\textbf{Robot self-recognition and self-other distinction} capabilities are studied through both non-learning and learning-based methods. 
Non-learning-based approaches involve correlating observed motion with robot actions using techniques such as mutual information~\cite{edsinger2006_mutual_self_recognition, yang2020morphology}, dense image Jacobian estimation~\cite{toshimitsu2022_dije_self_recognition}, or temporal delay~\cite{michel2004_motion_self_recognition}. 
Learning-based approaches are studied at both intermediate visual cue level, such as optical flow~\cite{lanillos2020robot_self_other} and pixel level~\cite{byravan2018_se3_pose_net, florence2020_robot_segmentation, sancaktar2020_pixel_body_perception}.
The training data of the learning-based approaches are self-labelled by associating the motor inputs with the observations. 
However, the development of self-recognition and self-other distinction through end-to-end visuomotor policy learning and their impact on the policies remains unexplored. 

\textbf{Image-proprioception integration} is fundamental for end-to-end visuomotor policy learning in robotics and plays a vital role in cultivating body ownership in humans.
While the specific integration mechanism in human cognition remains elusive, robotics leverage the \textit{concatenation paradigm} for this purpose.
Here, image representations are extracted from either single or sequential observations using CNNs~\cite{finn2016_spatial_encoder_vs, mandlekar2021_what_matters_robot_learning, chi2023_diffusion_policy}, or  transformers~\cite{xiao2022_masked_pretrain_control_vit, brohan2022_rt1}.
Simultaneously, the proprioceptive states, which may include manipulator joint positions, velocities, and end-effector poses, are directly read from the manipulator~\cite{xiao2022_masked_pretrain_control_vit, chi2023_diffusion_policy}, or further encoded through a multi-layer perceptron~\cite{zhuang2020_lyrn}.
The two representations are then concatenated for the downstream task. 
Despite this integration, end-to-end learning approaches have yet to demonstrate an innate development of self-recognition and self-other distinction capabilities.

\textbf{Contrastive learning} techniques form the cornerstone of RcP for tackling the challenge of developing body ownership with existing single-manipulator datasets. 
These techniques~\cite{oord2018representation_cpc, chen2020_simclr, laskin2020_curl, he2020_moco} employ discriminative learning objectives to encourage similarities within positive pairs and dissimilarities within negative pairs.
The generation of these pairs are typically constructed through sequences of data augmentation techniques~\cite{chen2020_simclr, laskin2020_curl, shen2022_un-mix}.
A typical training process involves dual encoders to process two sets of separately augmented samples~\cite{chen2020_simclr, verma2021_contrastive_learning_separate_encoders, wu2018_memorybank}. 
In this work, we adopt the Momentum Contrastive Learning framework (MOCO)~\cite{he2020_moco} proposed by He et al., a strategy where the second encoder's weights are updated as a momentum moving average of the first's, and maintaining a negative sample queue for past keys to increase the negative sample size.

\section{Methodology}

\subsection{The Robot-centric Pooling Module}
\label{subsec: RcP}

As depicted in \figref{subfig:RcP}, given an image-proprioception pair $(\boldsymbol{x}, \boldsymbol{p})$, the RcP module uses a cross-attention mechanism~\cite{vaswani2017attention} to align the image and proprioceptive latent representations while enabling self-supervised contrastive learning. It outputs a context vector $\boldsymbol{h}$ for contrastive learning and Image-Proprioception Alignment (IPA) scores. The IPA scores are then transformed into a binary mask, identifying the most relevant feature for the regression task (\figref{subfig:system_overview}).


The RcP image encoder $f_{\boldsymbol{\theta}}$, with learnable parameters $\boldsymbol{\theta}$, encodes an input image $\boldsymbol{x}$ into image keys $\boldsymbol{k}$ and values $\boldsymbol{v}$:
\begin{equation}
(\boldsymbol{k}, \boldsymbol{v}) = f_{\boldsymbol{\theta}}(\boldsymbol{x}).
\label{eq:rcp_im_encoder}
\end{equation}
Specifically, the RcP image encoder's backbone is a modified \texttt{ResNet18}~\cite{he2016_resnet} with the last average pooling and the classification layers removed.
Two learnable projection matrices, denoted as $W_k,\ W_v$, project the layer-normalised~\cite{ba2016_layernorm} features into image keys $\boldsymbol{k}$ and values $\boldsymbol{v}$:
\begin{align}
    \boldsymbol{k} &= W_k\left(\mathrm{LN}(\mathrm{ResNet}_{\boldsymbol{\psi}}(\boldsymbol{x}))\right) \in \mathbb{R}^{(hw) \times d},
        \label{eq:kv}
\\
    \boldsymbol{v} &= W_v\left(\mathrm{LN}(\mathrm{ResNet}_{\boldsymbol{\psi}}(\boldsymbol{x}))\right) \in \mathbb{R}^{(hw) \times d}.
\end{align}
Here, $\boldsymbol{\psi}$ denotes the learnable parameters, $\mathrm{LN}(\cdot)$ denotes the LayerNorm operation, \(hw\) represents the product of the spatial dimensions, and \(d\) specifies the feature dimension.

Frequency encoding has shown advantages in mapping continuous coordinate inputs to higher dimensions, which facilitates better approximation of high-frequency functions~\cite{mildenhall2021_nerf}.
Since robot proprioceptive states $\boldsymbol{p}$ are continuous and periodic (e.g., joint angles), they are first encoded into a frequency representation, formally defined as:
\begin{equation}
    \gamma(\boldsymbol{p}) = \bigoplus_{l=0}^{L-1} \left( \sin(2^l \pi \boldsymbol{p}), \cos(2^l \pi \boldsymbol{p}) \right),
\end{equation}
where $ \bigoplus $ denotes the concatenation of all frequency levels from $0$ to $L-1$. The resulting representation $\gamma(\boldsymbol{p})$ is then processed through a two-layer perceptron $Q(\cdot)$, yielding a proprioception query:
\begin{equation}
    \boldsymbol{q} := Q(\gamma(\boldsymbol{p})) \in \mathbb{R}^{1 \times d}.
    \label{eq:proprio_query}
\end{equation}
A proprioception decoder with mirrored architecture as $Q(\cdot)$ facilitates the learning of the proprioception query, forcing meaningful latent representation of the robot state. 

Denote the IPA scoring map as $s\in\R^{h\times w}$, and the vectorisation operator as $\mathrm{vec}:\R^{h\times w}\rightarrow \R^{1\times (hw)}$.
Based on the image keys \( \boldsymbol{k} \) and the proprioception query \( \boldsymbol{q} \), the IPA scores are computed as their cosine similarities: 
\begin{equation}
    \mathrm{vec}(\boldsymbol{s}) = \text{softmax}(\boldsymbol{q}\boldsymbol{k}^\top)
    \label{eq:ipa_score}
\end{equation}

Finally, the image values $\boldsymbol{v}\in\R^{(hw)\times d}$ are aggregated based on the IPA score vector to form the context vector $ \boldsymbol{h}$ following the cross-attention formulation:
\begin{equation}
    \boldsymbol{h} = \text{softmax}(\boldsymbol{q} \boldsymbol{k}^\top)\, \boldsymbol{v}.
    \label{eq:context_vec}
\end{equation}

In practice, the IPA scores between self-distractors that resemble the robot's state can be similar, introducing noise into the policy regression pipeline. 
To mitigate this, an \(\texttt{argmax}\) operation is applied to the IPA scoring map, generating a binary mask that highlights only the image region most relevant to the robot's state:
\begin{equation}
\mathrm{RcP}(\boldsymbol{x}, \boldsymbol{p}) := \mathbf{1}_{i j}\left({\underset{i,j}{\mathrm{argmax}} \, \boldsymbol{s}_{ij}}\right),
\label{eq:ipa_argmax}
\end{equation}
where the $\mathrm{argmax}$ operation identifies the indices $(i, j)$ corresponding to the maximum value in the IPA scoring map $\boldsymbol{s}$, and $\mathbf{1}_{ij}(\cdot)$ represents the indicator function that assigns 1 to the location $(i, j)$ and 0 elsewhere, resembling a global max-pooling operation with spatial locations specified by the maximum IPA score (Eq.~\eqref{eq:rcp}).

\begin{figure}
    \centering
    \subfloat[Contrastive Pipeline]{\includegraphics[height=0.22\textwidth]{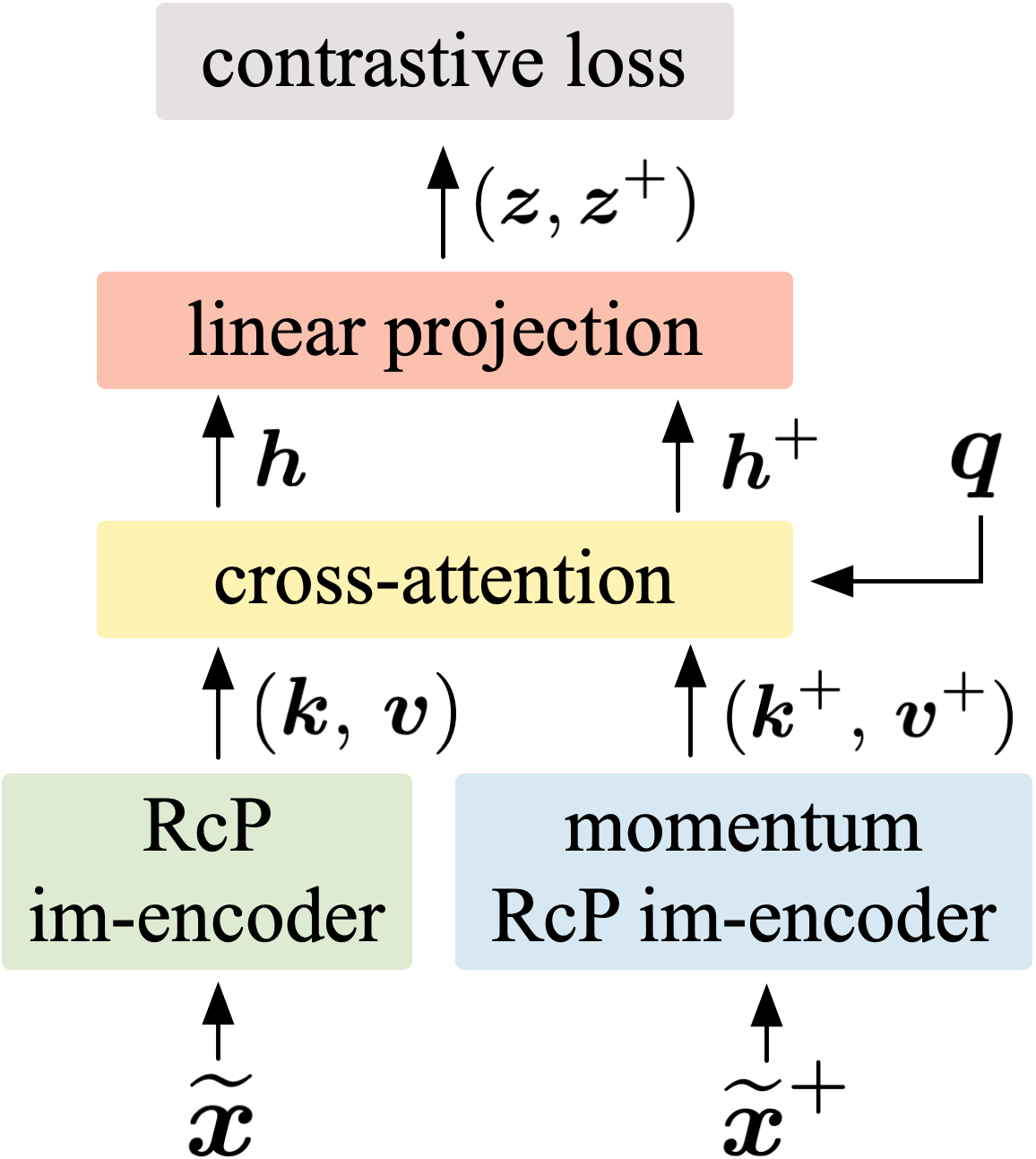} \label{subfig:moco}}
    \hfill
    \subfloat[Data Augmentation]{
        \includegraphics[height=0.22\textwidth]{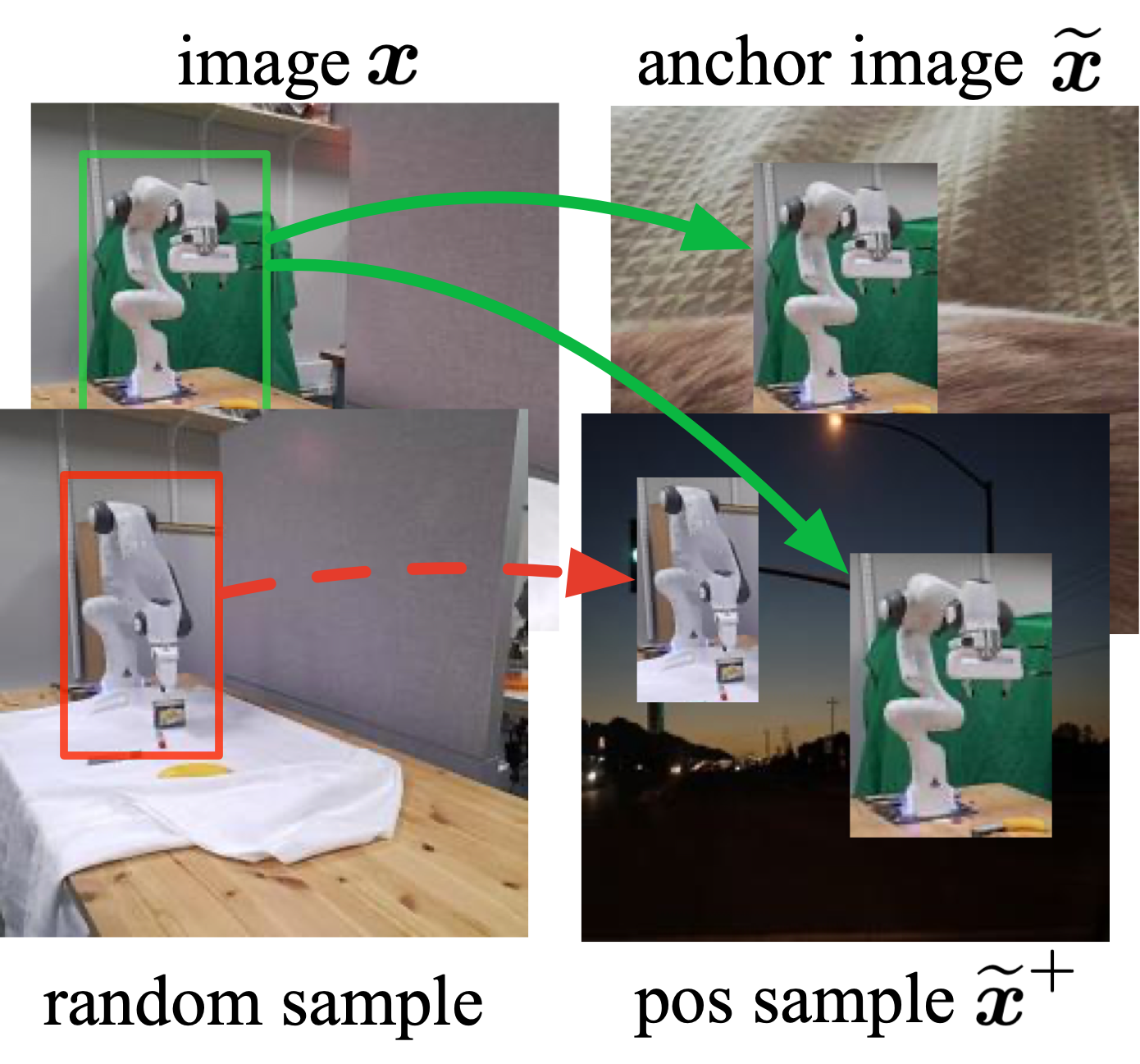} \label{subfig:data_augmentation}}
    \caption {\textbf{Illustration of the Contrastive Learning Framework.}
    \textbf{(a):} Similar to the pipeline proposed in MOCO~\cite{he2020_moco}, the augmented images are separately encoded by the RcP's image encoder and its momentum averaging copy. \textbf{(b):} For each image, the manipulator region is cropped and pasted onto two random backgrounds at random spatial locations (green firm arrows). A self-distractor is cropped from a random image drawn from the training dataset and randomly pasted onto one of the augmented images (red dashed arrow).}
    \vspace{-2mm}
\end{figure}

\subsection{Emergence of Body Ownership via Contrastive Learning}
\label{subsec:contrastive_learning}
The image keys corresponding to the `self' should achieve high IPA scores in comparison to those associated with the environment or other bodies.
Therefore, context features derived from the same proprioceptive query are expected to be similar, despite changes in background or self-distractors. While context features extracted from different queries should show clear dissimilarity.


RcP achieves this core capability through contrastive learning techniques in a self-supervised fashion. 
Firstly, a stochastic data augmentation module transforms an image $\boldsymbol{x}$ into an anchor image $\widetilde{\boldsymbol{x}}$ and the corresponding positive image $\widetilde{\boldsymbol{x}}^+$.
More specifically, as shown in~\figref{subfig:data_augmentation}, this process involves cropping the manipulator region from the image using a roughly calibrated camera and then pasting this crop onto two randomly selected backgrounds at varied spatial locations. 
Meanwhile, a self-distractor is cropped from a random image within the training dataset and pasted onto one of the two augmented images in a location that ensures the majority of the already pasted robot remains visible. 
Random scaling is applied to the crops and 
colour jittering as in~\cite{mandlekar2021_what_matters_robot_learning} is applied to the composed images.
Each of the two augmented images has an equal probability of being designated as either the anchor $\widetilde{\boldsymbol{x}}$ or the positive image $\widetilde{\boldsymbol{x}}^+$. 

As depicted in~\figref{subfig:moco}, similarly to MOCO~\cite{he2020_moco}, we employ two copies of the RcP image encoder (Eq.~\eqref{eq:rcp_im_encoder}) for contrastive learning.
Note that, the proprioception encoder and decoder are not copied.
The first copy $f_{\boldsymbol{\theta}}$ receives the gradient updates and the other $f_{\bar{\boldsymbol{\theta}}}$, referred to as the momentum image encoder, having its weights updated as a moving-average of $\boldsymbol{\theta}$:
\begin{equation*}
    \bar{\boldsymbol{\theta}} \leftarrow m\bar{{\boldsymbol{\theta}}} + (1-m) \boldsymbol{\theta},\ \text{where}\ m \in [0, 1).
\end{equation*}
Following Eq.~\eqref{eq:kv}, these encoders separately process the anchor $\widetilde{\boldsymbol{x}}$ and the positive image $\widetilde{\boldsymbol{x}}^+$ into the corresponding keys and values. Meanwhile, the proprioception state $\boldsymbol{p}$ is encoded into the proprioception query $\boldsymbol{q}$ via Eq.~\eqref{eq:proprio_query}.
The anchor and positive image-derived keys and values separately perform the cross-attention (as in Eq.~\eqref{eq:ipa_score}) operation with $\boldsymbol{q}$, producing the corresponding context vectors $\boldsymbol{h}$ and $\boldsymbol{h}^+$.


The context vector $\boldsymbol{h}^+$ encoded by the momentum copy updates the negative samples queue for the next training step.
Both context vectors undergo a shared linear projection before computing the contrastive objective.
Denote the learnable linear projection weights as $W\in\R^{d\times d}$, the context vector $\boldsymbol{h}$ after the linear projection is defined as $\boldsymbol{z}:= W \boldsymbol{h}$.
We use InfoNCE Loss~\cite{oord2018representation_cpc} as the contrastive objective.
Given $K$ projected context vectors saved in the negative sample queue, $\{\boldsymbol{z}^-_0,\, \dots,\,\boldsymbol{z}^-_K\}$, and the positive pair $(\boldsymbol{z},\, \boldsymbol{z}^+)$, the objective is expressed as:
\begin{equation}
    \mathcal{L}_\text{moco} = -\log\frac{\exp(\boldsymbol{z}\cdot\boldsymbol{z}^+/\tau)}{\sum_{i=0}^K\exp(\boldsymbol{z}\cdot \boldsymbol{z}^-_i/\tau)},
    \label{eq:contrastive_loss}
\end{equation}
where $\tau$ is a scalar temperature hyper-parameter.

\subsection{Policy Regression}
\label{subsec:policy_regression}
The image encoder for the policy regressor is also a modified version of \texttt{ResNet18}, with the last classification and average pooling layers removed. 
Two additional convolution layers with $3\times3$ kernels, denoted as $\mathrm{Conv}(\cdot)$, are added to enlarge the receptive field of extracted features.
This is critical to compensate for the receptive field reduction caused by the $\mathrm{argmax}$ operation within RcP, ensuring that the receptive field of features at each spatial location adequately covers the entire input image. 
The image feature after Robot-centric Pooling, \( \boldsymbol{g} \in \mathbb{R}^d \), is then expressed as
\begin{equation}
    \boldsymbol{g} = \mathrm{RcP}(\boldsymbol{x}, \boldsymbol{p})\otimes \mathrm{Conv}\left(\mathrm{ResNet}_{\boldsymbol{\phi}}(\boldsymbol{x})\right),
    \label{eq:rcp}
\end{equation}
where $\boldsymbol{\phi}$ denotes the learnable parameters within the image backbone, $\mathrm{RcP}(\cdot)$ represents the Robot-centric Pooling operation as defined in Eq.~\eqref{eq:ipa_argmax}, and \( \otimes \) denotes the element-wise multiplication broadcasting over the feature dimension.
The image feature $\boldsymbol{g}$ is then concatenated with the proprioceptive state \( \boldsymbol{p} \) to form the input vector for policy regression.

The policy regressor \( \pi_{\boldsymbol{\xi}}(\cdot) \) is a two-layer perceptron.
We use the L1 loss as the regression objective.
Given an observation \( (\boldsymbol{x}, \boldsymbol{p}) \) pair and the corresponding ground truth action \( \boldsymbol{a} \), the policy regression loss is 
\begin{equation}
    \mathcal{L}_\text{policy} = \lvert(\pi_{\boldsymbol{\xi}}(\boldsymbol{g}, \boldsymbol{p}) - \boldsymbol{a}\rvert.
\end{equation}
The Robot-centric Pooling module can be pre-trained or trained jointly with the regression pipeline.
When jointly trained, a preferential weighting $\lambda \in [0,\, 1)$ is applied to balance the gradients between the contrastive loss and the regression loss:
\begin{equation}
    \mathcal{L} = \mathcal{L}_{\text{policy}}  + \lambda( L_{\text{moco}}+\mathcal{L}_{\text{recon}}) ,
    \label{eq:loss}
\end{equation}
where $\mathcal{L}_{\text{recon}}$ is a mean square error loss for reconstructing the encoded proprioception query. 

\begin{figure*}
    \centering
    \subfloat[Randomised Distractors (Sim)]{
    \includegraphics[height=0.27\linewidth]{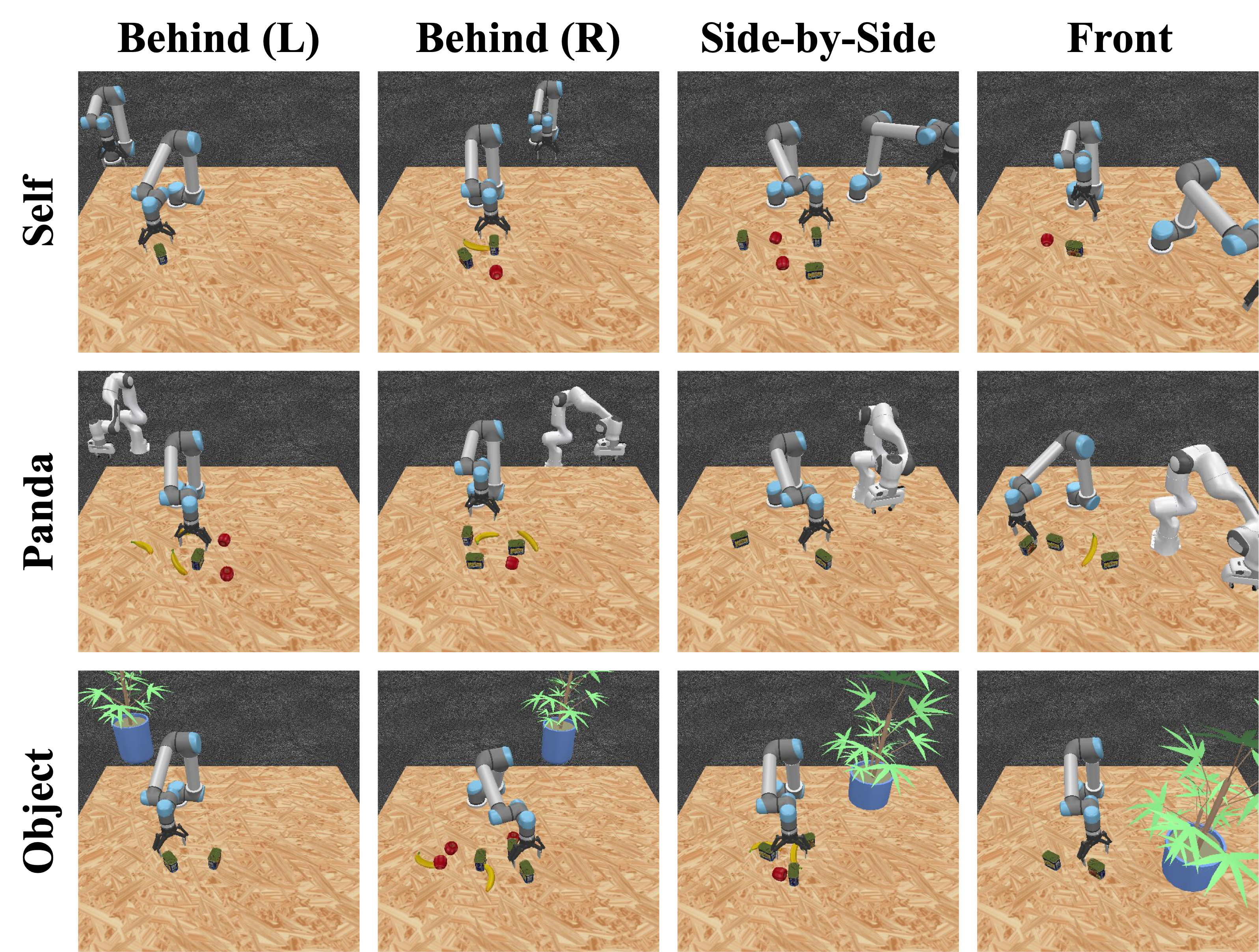} \label{subfig:sim_exps}
    }
    \hspace{2mm}
    \subfloat[Input Saliency Maps (Sim)]{    
    \includegraphics[height=0.27\linewidth]{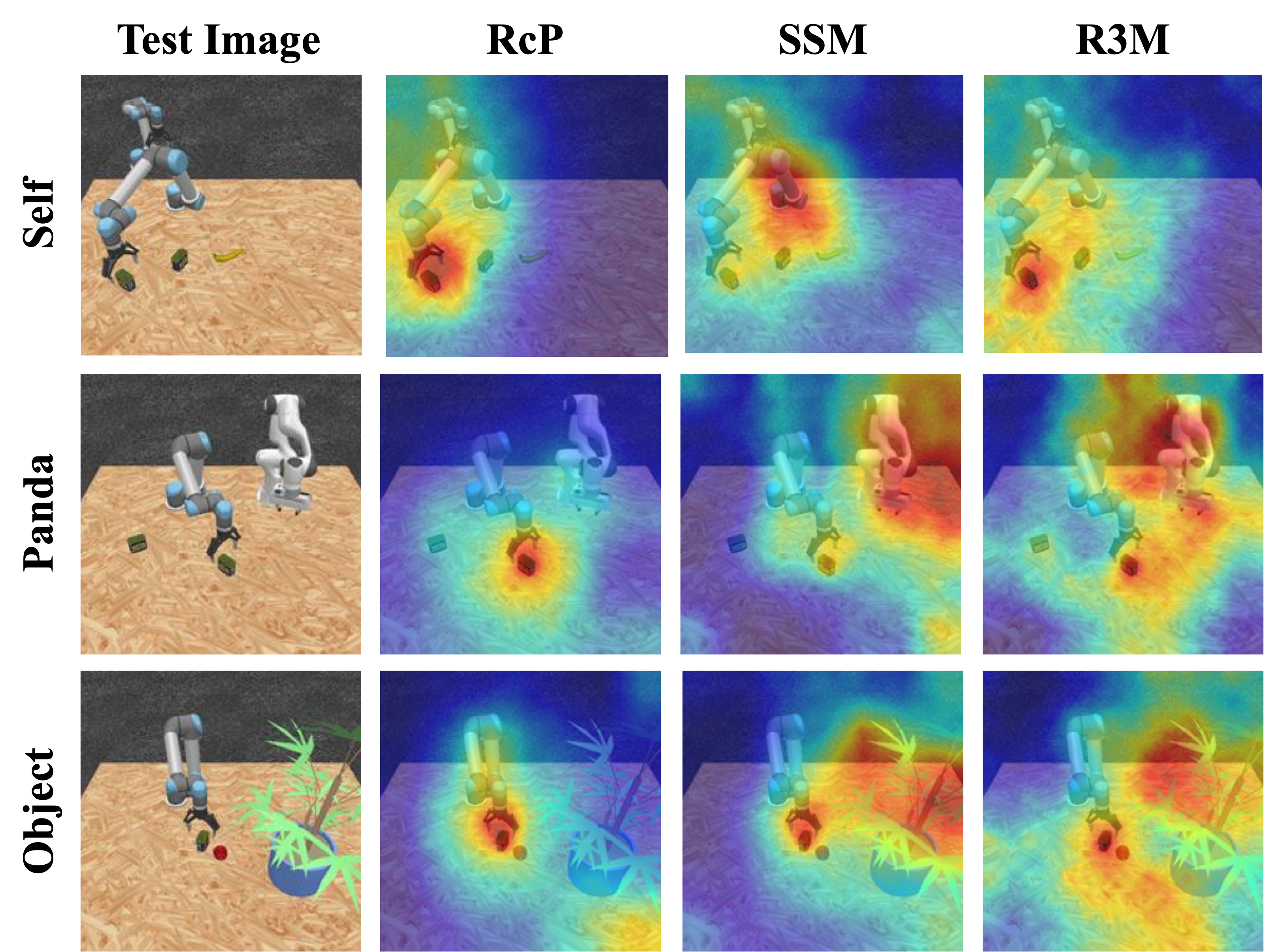}    \label{subfig:fullgrad_vis}
    }
    \hspace{2mm}
    \subfloat[Real Exps.]{    
    \includegraphics[height=0.27\linewidth]{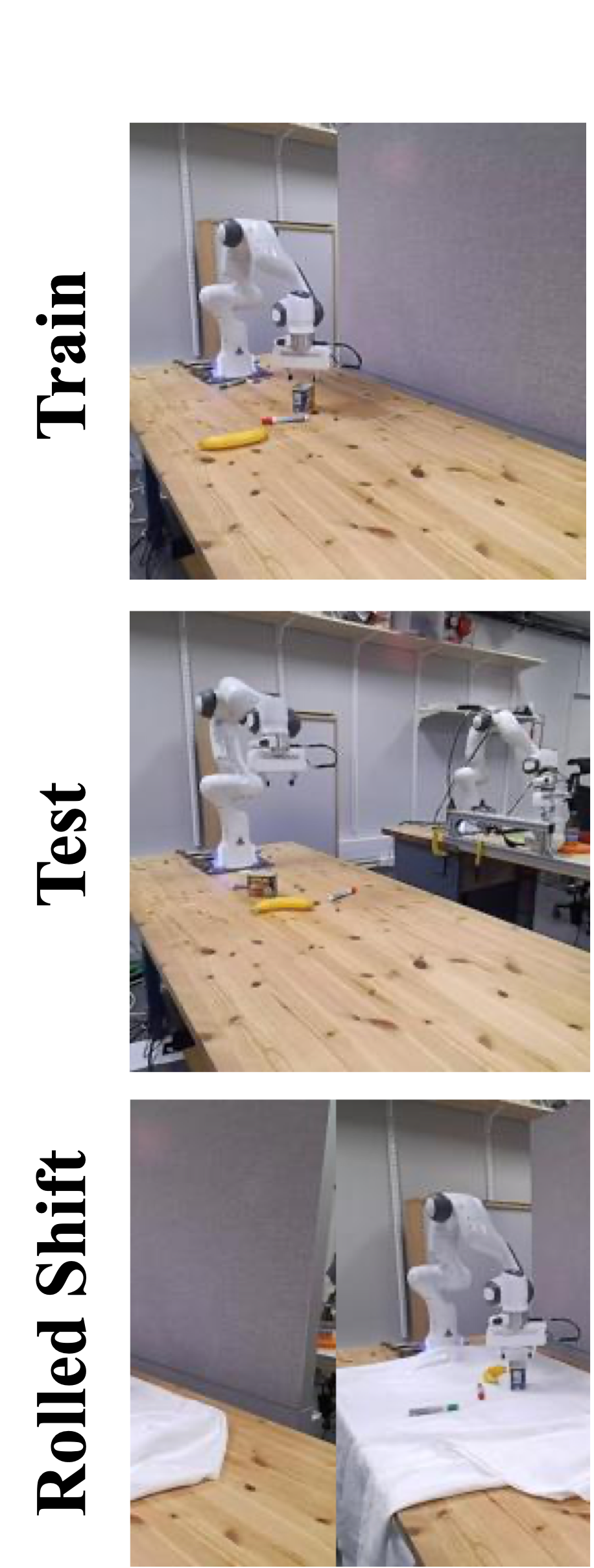}    \label{subfig:real_setup}
    }
    \caption{\textbf{Illustration of Simulated Experiments, Input Saliency Maps, and Real Experiments}. \textbf{(a):} Three distractors: the self-distractor, a Franka Panda robot, and a static object (a pot plant), are positioned at four distinct locations: behind the robot towards the left, behind the robot towards the right, alongside the robot, and in front of the robot. During the experiments, both the self-distractor and the Franka Panda execute random actions. \textbf{(b):} We employ an image saliency visualisation tool, FullGrad~\cite{srinivas2019fullgrad}, to visualise the activated image regions for different policies. \textit{RcP: Robot-centric Pooling, SSM: Spatial-Softmax}. \textbf{(c):} The real-world setup features a second-person camera view, with the robot dominant on the left side of the image. A movable divider can conceal and reveals the distractor robot against a less-structured background based on scenarios. The rolled-shift image is used for testing the networks' robustness against image shifts.
    }
\vspace{-3mm}
\end{figure*}

\section{Experiments}
To evaluate the effectiveness of Robot-centric Pooling (RcP) in fostering body ownership, we conduct a series of reaching experiments in both simulated and real-world environments, intentionally introducing distractions from the environment and other robot entities.
In all scenarios, we use second-person camera configurations.
This increases the system's susceptibility to distractors, providing us with a unique opportunity to assess the importance of body ownership in enhancing the robustness of the policies.

\subsection{Baselines and training details}
\label{subsec:training_details}
We select two standard Resnet18-based behaviour cloning networks as the baselines, distinguished by their pooling and pre-training methods. 
One is pre-trained on ImageNet1K~\cite{deng2009_imagenet}, with Spatial-Softmax (SSM) pooling~\cite{finn2016_spatial_encoder_vs}.
Unlike average pooling, Spatial-Softmax focuses on the 2D spatial locations of the highest activations within image feature maps.
It is used as one of the vision baselines in robomimic~\cite{mandlekar2021_what_matters_robot_learning} and the image-backbone in diffusion policy~\cite{chi2023_diffusion_policy}.
The second baseline, R3M~\cite{nair2022_r3m}, benefits from pre-training on the first-person human activity dataset Ego4D~\cite{grauman2022_ego4d} through a contrastive learning approach. 
While keeping the average pooling layer, R3M is reported to achieve more generalisable feature extraction for manipulation tasks.

Both baseline networks and the regression pipeline of our proposed method follow the same architectural framework as depicted in~\figref{subfig:system_overview}. 
All networks are trained using the AdamW optimiser~\cite{loshchilov2017_adamw}, with a learning rate of $1\mathrm{e}^{-4}$, weight decay of $1\mathrm{e}^{-6}$, and a mini-batch size of 128, over the course of 150 epochs. 
Input images are resized to $224\times 224$ pixels and normalised according to the corresponding pre-training scheme utilised. 
Random colour jittering and pixel shift (up to 7\% of the image size) are also applied during training. 
The proprioceptive state includes the manipulator's joint angles and the end-effector pose.
Specifically, the end-effector's pose is represented with a 3D translation component and a 6D representation~\cite{zhou2019_6d_rot} for the rotation.
The actions are the end-effector 6 Dof velocities.

Regarding the proposed method, the negative sample queue size, momentum $m$ and temperature $\tau$ for the contrastive pipeline are set to 4096, 0.95, and 0.1, respectively, and the preferential weight $\lambda$ in Eq.\eqref{eq:loss} is set to 0.05.
Both the RcP image encoder and the regression image encoder are initialised from weights pre-trained on ImageNet1K~\cite{deng2009_imagenet}.
To enhance the training efficiency of the contrastive pipeline, a warm-up phase is employed.
During this phase, slightly shifted versions of the image $\boldsymbol{x}$ are used for the initial 5 epochs before transitioning to training with strongly augmented sample pairs as detailed in~\secref{subsec:contrastive_learning}.

\subsection{Simulated Experiments}
The simulation environment is created in CoppeliaSim with PyRep~\cite{james2019_pyrep}. 
We use a UR5 as the robot manipulator, with up to three Spam Cans as the target instances and up to five non-target random objects (such as bananas, mugs etc).
These objects are randomly positioned and re-oriented within the $0.3\times0.5\,\text{m}^2$ workspace.
The end-effector's translation is randomised within a $0.4\times 0.2\times 0.2 \text{m}^3$ cuboid. 
Meanwhile, its rotation is adjusted within a downward-pointing cone. 
Overall, the dataset comprises 3000 trajectories, averaging 20 data points per trajectory.
We adopt the multi-instance reaching trajectory generation formulation as described in~\cite{zhuang2020_lyrn}.
Convergence itself is defined as the pose at which the velocity change falls below a predetermined threshold.
A reach is deemed successful when the tool-central point is located within a $3\, \mathrm{cm}^3$ cuboid centred at the target pose, and the deviation in the yaw angle is less than $10^\circ$.
Notably, within the workspace's central area for this evaluation, a one-pixel displacement corresponds to a displacement of $1.12\,\mathrm{cm}$.

\subsubsection{Robustness against Distractors}
As illustrated in~\figref{subfig:sim_exps}, we evaluate the emergence of body ownership by introducing various out-of-distribution distractors into the scene: a self-distractor, a Franka Panda robot, and a static sizable object, such as a pot plant. 
We position each distractor in four distinct locations relative to the robot: behind it towards the left (with partial visual overlap), behind it towards the right, alongside the robot, and directly in front of the robot. 
During tests, both the self-distractor and the Franka undertake random actions. 
The performance of each network is evaluated by averaging the outcomes across three different random seeds, with each instantiation of the network executing 50 trajectories.

Without distractors, Robot-centric Pooling, Spatial-Softmax and R3M achieve $90.0\%$, $86.0\%$ and $71.0\%$ reaching success rate, respectively.
Under the presence of distractors, as tabulated in Tab.~\ref{tab:sim_exp_results}, Robot-centric Pooling (RcP) demonstrated superior performance across various settings, outperforming Spatial-Softmax (SSM) and R3M in its ability to handle different types of unseen distractors at various locations.
Specifically, RcP achieved an average success rate of 76.2\% against self-distractors, 70.3\% when faced with the Franka Panda robot, and 82.7\% against a static plant distractor.
On the other hand, the SSM baseline showed variability, with its highest success rate being 62.7\% against the Franka Panda when positioned at the back left but fails entirely when the distractors are in the front position.


We also employ an image saliency visualisation tool, FullGrad~\cite{srinivas2019fullgrad}, for analysing the varying performance among different pooling methods.
The saliency maps in ~\figref{subfig:fullgrad_vis} from the Robot-centric Pooling (RcP) model exhibit focused attention on the robot and its target, emphasising RcP's effectiveness in isolating and utilising relevant features for regression tasks. 
In contrast, saliency maps from Spatial-Softmax and R3M models show less discrimination, erroneously blending features from both the robot and nearby distractors into regression.

As shown in \figref{subfig:fullgrad_vis}, when the self-distractor overlaps with the robot, the RcP network may take parts of the distractor for policy regression, hence the relatively lower success rate compared to other distractor locations.
This issue is hypothesised to stem from the inherent resolution constraints at the final feature map level, where a $224\times224$ input image is reduced to a $7\times7$ feature map by \texttt{ResNet18}, potentially insufficient for disentangling overlapping features with high precision. 
We leave further investigation for future work.
\setlength{\tabcolsep}{3pt}
\begin{table}
    \caption{\textbf{Reaching Success Rate $(\%)$ against differnt types of distractors at varying locations}. \textit{RcP: Robot-centric Pooling, SSM: Spatial-Softmax}.}
    \centering
    \begin{tabular}{ccccccc}
       \toprule
       & & \textbf{Back(L)} &  \textbf{Back(R)} &  \textbf{Side} &  \textbf{Front} & \textbf{Average}\\
        \midrule
        \multirow{3}{*}{\textbf{\rotatebox[origin=c]{90}{Self}}}  & \textbf{RcP} & \textbf{62.7\,$\pm$\,8.2} & \textbf{86.7\,$\pm$\,1.7} & \textbf{81.3\,$\pm$\,2.5} & \textbf{74.0\,$\pm$\,0.8} & \textbf{76.2\,$\pm$\,12.6} \\
             & \textbf{SSM} & 57.3\,$\pm$\,6.5 & 50.7\,$\pm$\,7.6 & 2.7\,$\pm$\,1.9 & 0.0\,$\pm$\,0.0 & 27.7\,$\pm$\,28.4 \\
             & \textbf{R3M} & 7.3\,$\pm$\,2.1 & 0.0\,$\pm$\,0.0 & 0.0\,$\pm$\,0.0 & 0.0\,$\pm$\,0.0 & 1.8\,$\pm$\,3.8 \\
            
        \midrule
        \multirow{3}{*}{\rotatebox[origin=c]{90}{\textbf{Franka}}}
         & \textbf{RcP} & 56.0\,$\pm$\,10.2 & \textbf{78.0\,$\pm$\,2.2} & \textbf{80.0\,$\pm$\,4.5} & \textbf{67.3\,$\pm$\,5.2} & \textbf{70.3\,$\pm$\,15.7} \\
         & \textbf{SSM} & \textbf{62.7\,$\pm$\,7.4} & 17.3\,$\pm$\,10.9 & 0.0\,$\pm$\,0.0 & 0.0\,$\pm$\,0.0 & 20.0\,$\pm$\,28.8 \\
         & \textbf{R3M} & 12.0\,$\pm$\,5.1 & 8.0\,$\pm$\,4.3 & 0.0\,$\pm$\,0.0 & 0.0\,$\pm$\,0.0 & 5.0\,$\pm$\,8.5 \\

        \midrule
        \multirow{3}{*}{\rotatebox[origin=c]{90}{\textbf{Plant}}}  & \textbf{RcP} & \textbf{82.0\,$\pm$\,2.9} & \textbf{85.3\,$\pm$\,2.1} & \textbf{82.0\,$\pm$\,2.9} & \textbf{81.3\,$\pm$\,0.9} & \textbf{ 82.7\,$\pm$\,5.0} \\
 & \textbf{SSM} & 78.0\,$\pm$\,1.4 & 84.7\,$\pm$\,2.1 & 29.3\,$\pm$\,8.7 & 0.0\,$\pm$\,0.0 & 48.0\,$\pm$\,36.1 \\
 & \textbf{R3M} & 49.3\,$\pm$\,4.0 & 6.7\,$\pm$\,1.2 & 16.7\,$\pm$\,6.2 & 0.0\,$\pm$\,0.0 & 18.2\,$\pm$\,20.4 \\
    \midrule
    \multirow{3}{*}{\rotatebox[origin=c]{90}{\textbf{Total}}}
         & \textbf{RcP} & \textbf{66.9\,$\pm$\,9.5} & \textbf{83.3\,$\pm$\,2.7} & \textbf{81.1\,$\pm$\,3.5} & \textbf{74.2\,$\pm$\,4.2} & \textbf{76.4\,$\pm$\,13.0} \\
         & \textbf{SSM} & 66.0\,$\pm$\,7.2 & 50.9\,$\pm$\,15.8 & 10.7\,$\pm$\,8.4 & 0.0\,$\pm$\,0.0 & 31.9\,$\pm$\,33.5 \\
         & \textbf{R3M} & 22.9\,$\pm$\,10.2 & 4.9\,$\pm$\,3.1 & 5.6\,$\pm$\,5.3 & 0.0\,$\pm$\,0.0 & 8.3\,$\pm$\,14.7 \\
    \bottomrule
    \end{tabular}
    \label{tab:sim_exp_results}
    \vspace{-5mm}
\end{table}

\begin{figure*}
    \centering
    \subfloat[Translation Error After Self-Distractor's Presence]{
    \includegraphics[height=0.27\textwidth]{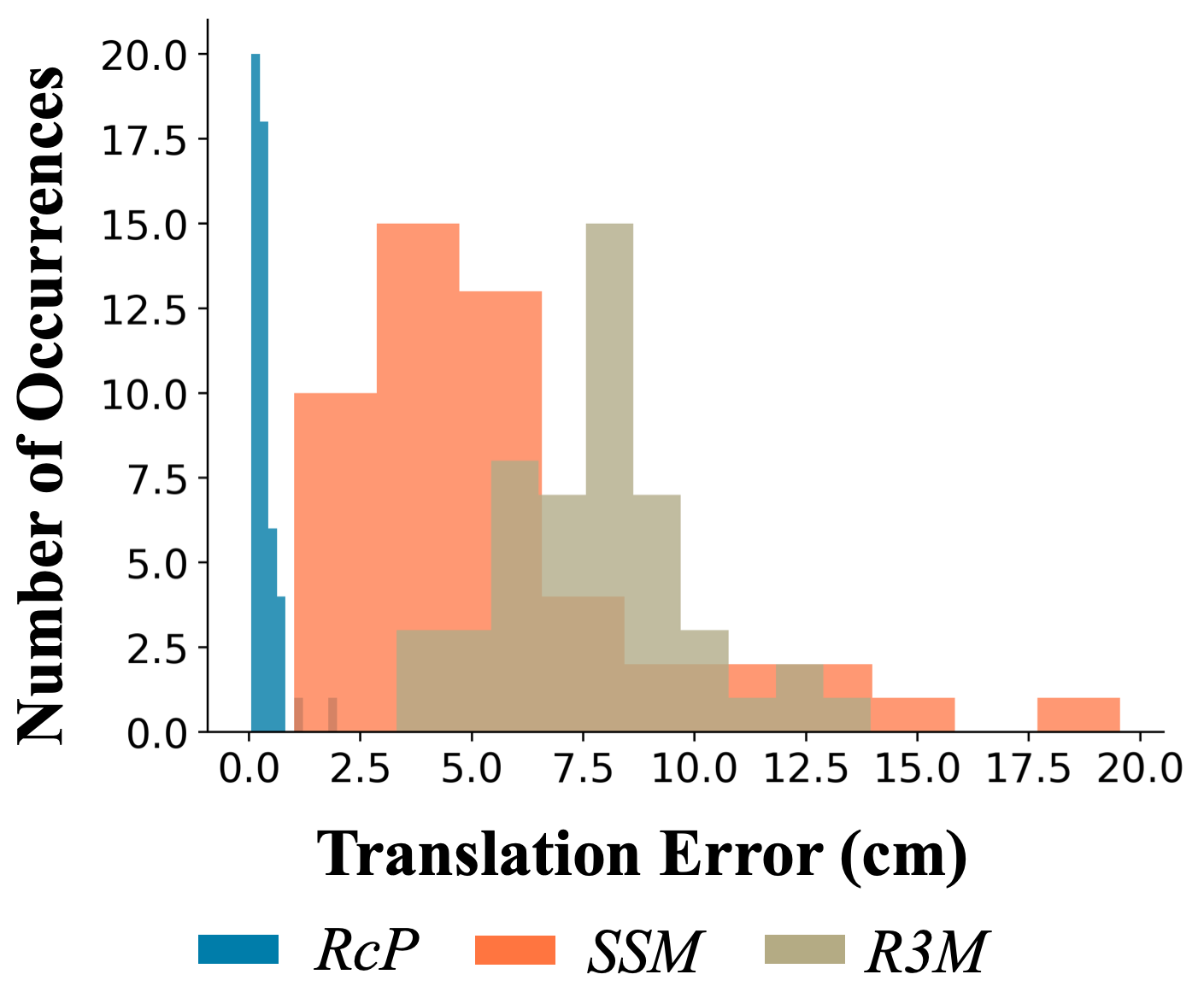}
    \label{fig:translation_error}
    }
    \hspace{3mm}
    \subfloat[Reaching success vs percentage of pixel shift.]{\includegraphics[height=0.27\textwidth]{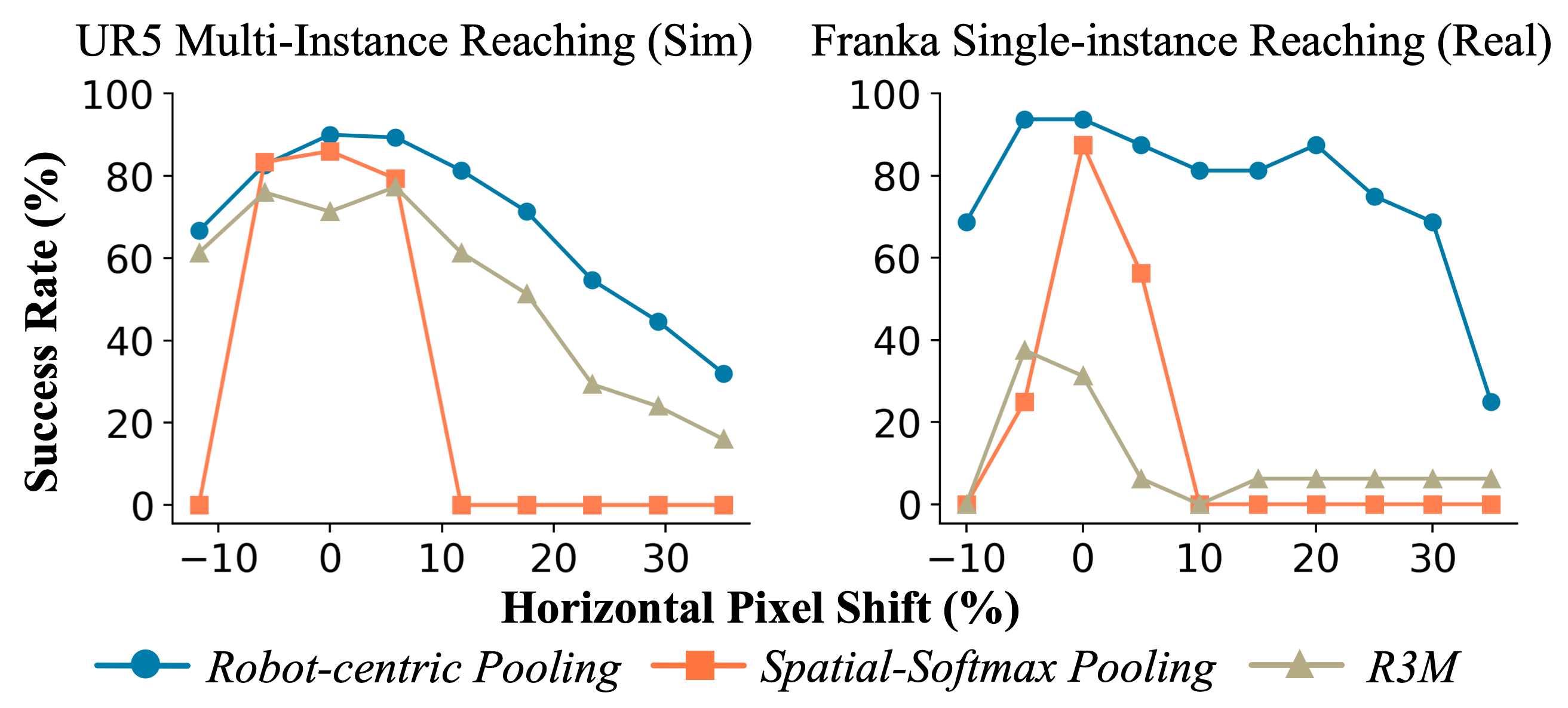} 
    \label{fig:pixel_shift}
    }
    \caption{\textbf{(a): The Translation Error after Self-Distractor's Presence (50  Target Poses).} The translation error is measured at the robot's tool-central point upon convergence, comparing trajectories aimed at the same target with and without a self-distractor. 
    \textbf{(b): Reaching success vs percentage of pixel shift.} UR5 multi-instance reaching experiment (Left) and Franka single-instance reaching task (Right).
    }
\vspace{-3mm}
\end{figure*}

\subsection{Real-World Experiments}

As shown in ~\figref{subfig:real_setup}, the real-world setting features a side second-person camera view, with the robot dominant on the left side of the image.
A movable divider conceals and reveals the self-distractor robot against a less-structured background.
During the reaching tasks, the robot starts from a predefined home position, targeting a Spam Can that is randomly positioned and reoriented within a designated area of $0.3\times0.3\mathrm{m}^2$. 
We collect 200 real-world trajectories combined with 1000 simulated trajectories for training.
The networks are first trained with the mixed datasets and then fine-tuned based on real-world data.
The training protocol is identical to the procedure described in~\secref{subsec:training_details}.
Each network is selected out of three random seeds, based on their reaching performance (without distractors) in the simulator.  

For evaluating the reaching success and the effectiveness of RcP's robustness against distractions, we position the target at 50 random locations. For each target location, we collect 6 reaching trajectories in total, one reaching attempt from each network (RcP, SSM, and R3M), with and without revealing the self-distractor.
A trajectory is deemed successful if the robot can establish a firm grasp on the target by extending 3 cm downward and then closing the gripper. 

Without the presence of the self-distractor, the reaching success rate is 96.0\%, 90.0\%, and 40.0\% for RcP, SSM and R3M, respectively. 
When introducing the self-distractor, RcP demonstrates remarkable robustness, with only a 4\% decrease in its success rate.
On the other hand, both SSM and R3M exhibit dramatic declines in performance, plummeting to 14\% and 2\%, respectively.

The histogram presented in \figref{fig:translation_error} outlines the deviation in translation at the robot's tool-central point upon convergence, comparing trajectories aimed at the same target with and without the presence of a self-distractor. 
For Robot-centric Pooling (RcP), deviations largely stay within a compact 1cm range, highlighting RcP's consistency. 
On the other hand, Spatial-Softmax and R3M baselines exhibit significantly higher mean errors and wider error distributions, underscoring the pronounced negative impact of the distractor on their performance.

\subsection{Robustness against Pixel Shift}


Through both simulated UR5 multi-instance reaching tasks and real-world Franka Panda single-instance reaching tasks, the proposed Robot-centric Pooling (RcP) method demonstrates consistently superior resilience against aggressive horizontal pixel shifts compared to the baseline method. 
In both the simulation and real-world experiments, the input images are subject to rolled shifts ranging between -10\% and +35\%, with increments of 5\%. 
The rolled-shift operation wraps the pixels extending beyond one edge of the image back onto the opposite edge (\figref{subfig:real_setup}).
In the simulation, for each shift increment, the performance of each network is averaged across three different random seeds, with each instantiation of the network executing 50 trajectories.
The target's pose shuffled for each trajectory.
In the real-world experimental setup, the target's pose undergoes 16 random shuffles. 
For each of these poses, every network carries out one reaching attempt for each shift increment,

As shown in \figref{fig:pixel_shift}, RcP maintains a gradually declining success rate in both simulated and real-world experiments, yet significantly outperforms baseline lines.
We attribute this inherent robustness against pixel shift to RcP's self-referential nature, which directs the focus of perception from a broad, global view to a localised, robot-centric perspective. 
In the simulation, R3M exhibits a lower yet similar performance curve as RcP.
This resemblance is likely due to the rolled image shift's characteristic of maintaining the overall pixel intensity distribution, which aligns well with the nature of R3M's global average pooling method.
However, R3M exhibits high sensitivity to the sim-to-real domain gap. 
Spatial-Softmax demonstrates a remarkably narrow peak in its performance curve, matching the 7\% random pixel shift parameter set during training. 
This observation suggests its limited capacity for generalisation across varying degrees of pixel shift, indicating the development of strong spatial biases.
%
%
\section{Conclusion}
In this work, we explored the concept of body ownership within the context of end-to-end visuomotor policy learning. 
We showed that the conventional end-to-end learning models do not spontaneously develop a sense of body ownership and are highly sensitive to distractors.
We introduced Robot-centric Pooling (RcP) that aggregates image features based on image-proprioception alignment. 
We demonstrated that replacing the conventional final pooling layer with RcP allows the learned policy to develop pronounced self-recognition and self-other distinction capabilities. 
Notably, tested with reaching tasks, in both simulated and real-world settings, the policy equipped with RcP exhibited its strong robustness against environmental distractions and self-distractors, significantly surpassing the conventional baselines. 
Furthermore, benefiting from the robot-centric nature of RcP, the learned policy exhibited enhanced resilience to aggressive image shifts.

In the proposed Robot-centric Pooling, we primarily focused on exploring the spatial alignment aspect for gaining body ownership capability. 
The potential performance gains by including observational history and robot dynamics have not yet been explored. 
It entails not only spatial but also temporal alignment, i.e., aligning both `seen' and `felt' positions, velocities, and accelerations, introducing the addition of visuomotor alignment. 
We leave this exciting direction for future work.
\section{Acknowledgement}
This work has been supported through WASP - Wallenberg AI, Autonomous systems and Software Program.

%


\bibliographystyle{util/IEEEtran}
\bibliography{ref_lib}

\end{document}